\documentclass[dvipsnames]{article}

\usepackage{microtype}
\usepackage{graphicx}
\usepackage{subfigure}
\usepackage{booktabs} 

\usepackage[accepted]{icml2024}

\usepackage{amsmath}
\usepackage{amssymb}
\usepackage{mathtools}
\usepackage{amsthm}

\usepackage{xcolor}         

\usepackage[colorlinks, linkcolor=RoyalBlue, anchorcolor=BrickRed, citecolor=RoyalBlue, urlcolor=RoyalBlue]{hyperref}

\usepackage[capitalize,noabbrev]{cleveref}

\crefname{section}{§}{§§}
\Crefname{section}{§}{§§}

\newcommand\refsec[1]{\hyperref[sec:#1]{§\ref{#1}}}

\newcommand\takeaway[1]{\hypertarget{#1}{\textcolor{RoyalBlue}{\underline{\textsc{#1}}}}}

\theoremstyle{plain}
\newtheorem{theorem}{Theorem}[section]

\theoremstyle{definition}

\newtheorem{assumption}[theorem]{Assumption}
\theoremstyle{remark}

\usepackage[textsize=tiny]{todonotes}

\icmltitlerunning{Unified View of Grokking, Double Descent and Emergent Abilities: A Perspective from Circuits Competition}

\begin{document}

\twocolumn[
\icmltitle{Unified View of Grokking, Double Descent and Emergent Abilities: A Perspective from Circuits Competition}



\icmlsetsymbol{equal}{*}

\begin{icmlauthorlist}
\icmlauthor{Yufei Huang}{tsinghua}
\icmlauthor{Shengding Hu}{tsinghua}
\icmlauthor{Xu Han}{tsinghua}
\icmlauthor{Zhiyuan Liu}{tsinghua}
\icmlauthor{Maosong Sun$^{\dag}$}{tsinghua}
\end{icmlauthorlist}

\icmlaffiliation{tsinghua}{Tsinghua University, Beijing, China}

\icmlcorrespondingauthor{Maosong Sun}{sms@mail.tsinghua.edu.cn}

\icmlkeywords{Deep Learning, Grokking, Double Descent, Emergent Abilities}

\vskip 0.3in
]



\printAffiliationsAndNotice{}  

\begin{abstract}
Recent studies have uncovered intriguing phenomena in deep learning, such as \textit{grokking}, \textit{double descent}, and \textit{emergent abilities} in large language models, which challenge human intuition and are crucial for a deeper understanding of neural models. In this paper, we present a comprehensive framework that provides a unified view of these three phenomena, focusing on the competition between memorization and generalization circuits. This approach, initially employed to explain \textit{grokking}, is extended in our work to encompass a wider range of model sizes and training data volumes. Our framework delineates four distinct training dynamics, each depending on varying combinations of model size and training data quantity. Utilizing this framework, we provide a detailed analysis of the \textit{double descent} phenomenon and propose two verifiable predictions regarding its occurrence, both substantiated by our experimental results. Moreover, we expand our framework to the multi-task learning paradigm, demonstrating how algorithm tasks can be turned into emergent abilities. This offers a novel perspective to understand \textit{emergent abilities} in Large Language Models.
\end{abstract}

\section{Introduction}
\label{introduction}

\begin{figure}[t]
    \centering
    \includegraphics[width=0.95\linewidth]{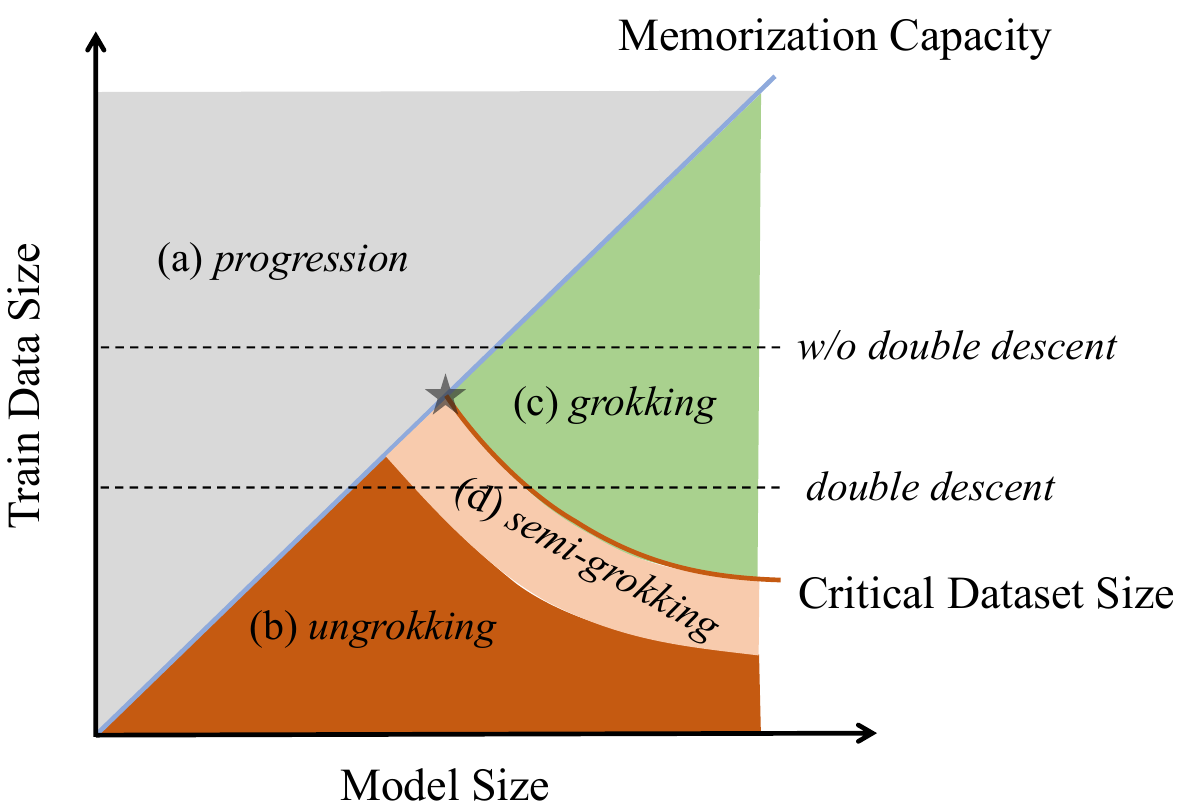}
    \caption{The increasing memorization capacity and decreasing critical dataset size for larger models split the figure into four distinct zones including \textit{progression}, \textit{ungrokking}, \textit{grokking} and \textit{semi-grokking}. Each zone will show a specific training dynamic.}
    \label{fig:theory_graph}
\end{figure}

There are several interesting phenomenons in Deep Learning, among which \textit{grokking}~\citep{DBLP:journals/corr/abs-2201-02177}, \textit{double descent}~\citep{DBLP:conf/iclr/NakkiranKBYBS20} and \textit{emergent abilities}~\citep{wei2022emergent} in current Large Language Models attract a lot of attention. Understanding these phenomena is important for us to reveal the mechanism of deep learning. Plenty of works~\citep{DBLP:conf/nips/LiuKNMTW22, DBLP:conf/iclr/LiuMT23, DBLP:journals/corr/abs-2206-04817, DBLP:journals/corr/abs-2309-02390, schaeffer2023are, michaud2023the} have been done to explain these phenomenons from different perspectives. However, these works all concentrate on a single phenomenon and explain them separately. In this work, we provide a preliminary study to give a unified view of these three phenomena from the perspective of competition between memorization circuits and generalization circuits in neural models.


\begin{figure*}[t]
    \centering
    \includegraphics[width=1\linewidth]{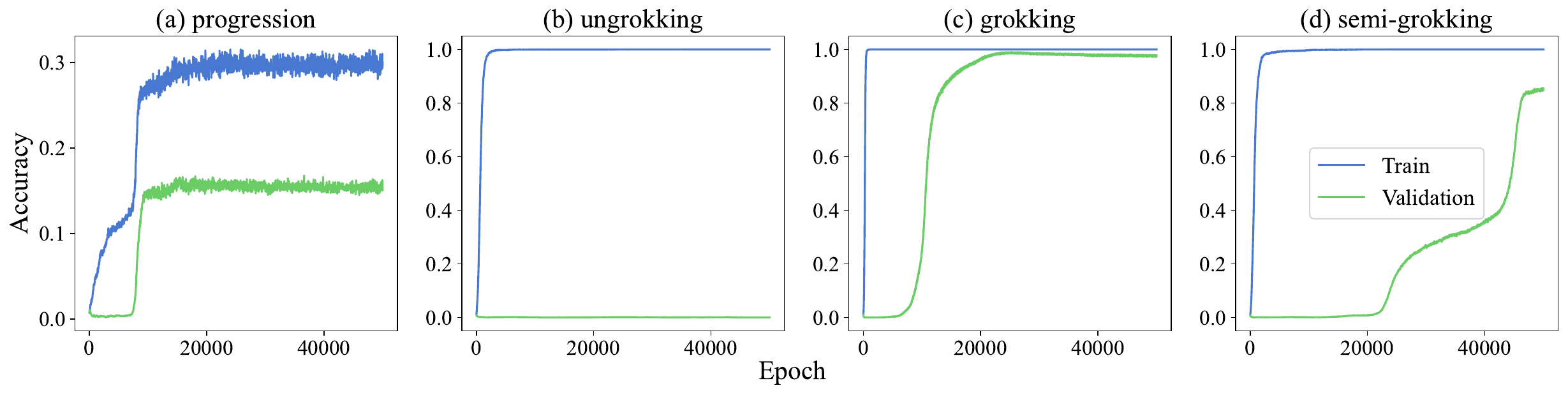}
    \caption{This figure illustrates the four distinct training dynamics that correspond to the zones identified in Figure~\ref{fig:theory_graph} and discussed in Section~\ref{sec:proposed_framework}. Each panel represents a specific dynamic: (a) \textit{Progression}, demonstrated using a model with a hidden size of $8$ and trained on $3000$ data points. (b) \textit{Ungrokking}, shown with a model having a hidden size of $32$, trained on $2600$ data points. (c) \textit{Grokking}, visualized using a model with a larger hidden size of $64$, also trained on $3000$ data points. These dynamics exemplify the variable responses of models with different configurations to specific training data volumes. (d) \textit{Semi-Grokking}, depicted with a model of hidden size $32$, trained on $3000$ data points.}
    \label{fig:four_zones}
\end{figure*}

Our work is based on \citet{DBLP:journals/corr/abs-2309-02390}'s explanation for \textit{grokking}. They attribute \textit{grokking} to the competition between two distinct types of circuits in the model: one responsible for memorization, which achieves high training accuracy but poor validation accuracy, and another for generalization, capable of high performance in both training and validation. The latter, although slower to develop, proves more efficient in terms of parameter norms, leading to the model finally transferring from memorization to generalization to achieve higher efficiency. Intriguingly, the efficiency of the memorization circuit is inversely proportional to the volume of training data, indicating that larger datasets reduce their efficiency. In contrast, the efficiency of the generalization circuit remains consistently stable, regardless of the size of the training data. Consequently, \citet{DBLP:journals/corr/abs-2309-02390} established a critical dataset size $D_{crit}$, a specific range within which memorization and generalization circuits exhibit comparable efficiency, and beyond which grokking is likely to occur.

Building upon this theory, our study expands the investigation to various model sizes. We firstly observe that smaller models require a larger critical dataset size for grokking, implying an inverse relationship between model size and the necessary amount of training data for grokking. Conversely, a model's memorization capacity is directly proportional to its size, indicating that smaller models have a reduced capacity for memorization. We confirm these relationships with plenty of experiments in Section~\ref{sec:study_about_grokking}.

The reverse relationships with model size inevitably lead to an intersection point of the two curves (black star in Figure~\ref{fig:theory_graph}). At this intersection, the model's memorization limit is reached, and the efficacy of this extreme memorization circuit is comparable to its generalization circuit. Besides, the two curves create four distinct zones on the graph, each reflecting a unique training dynamic in our experiments as shown in Figure~\ref{fig:four_zones}. (a) Excess training data or reduced model size hinders complete memorization of training data, causing model to exhibit an interesting phenomenon where model first memorizes part of the training data with zero validation performance and then generalizes to part of validation data with an increase in train accuracy either, which we name it \textit{progression}. (b) Insufficient data boosts memorization efficiency over generalization, causing model to choose pure memorization without generalization, which is consistent with \textit{ungrokking} stated by \citet{DBLP:journals/corr/abs-2309-02390}. (c) Conversely, increasing model size diminishes memorization efficiency relative to generalization, causing model transfer from memorization to generalization after training enough steps, which is exactly \textit{grokking}~\citep{DBLP:journals/corr/abs-2201-02177}. (d) When the number of training data points approximates the critical dataset size, the model exhibits a phenomenon named \textit{semi-grokking}, characterized by moderate generalization capabilities. This behaviour was first identified by \citet{DBLP:journals/corr/abs-2309-02390}.

Analyzing the depicted graph, we anticipate that the function representing final validation performance with model size will exhibit varying trends, contingent upon the quantity of training data. Specifically, when the quantity of training data surpasses the intersection point, an increase in model size leads the model through \textit{progression} to \textit{grokking}, resulting in a consistently positive correlation between model size and final validation performance. In contrast, for training data volumes below this point, the model undergoes \textit{progression}, \textit{ungrokking}, \textit{semi-grokking} and then \textit{grokking}, creating a function that first increases, then decreases, and ultimately increases again with increasing model size. This pattern exemplifies the \textit{double descent} phenomenon~\citep{Belkin_2019, DBLP:conf/iclr/NakkiranKBYBS20}. Furthermore, based on our analysis, we conduct experiments to transform a validation performance function of model size, which initially does not demonstrate clear \textit{double descent}, into one with obvious \textit{double descent}. This transformation is achieved by shifting the critical dataset curve upward, thereby also shifting the intersection point towards the upper right.

Further, we extend our experiments to the multitask learning paradigm where an algorithm task and a pure memorization task are mixed to train the model. Interestingly, adding a pure memorization task largely hinders model from formalising the generalization circuits for the algorithm task. With model size increasing, model always achieves near zero validation performance until a relatively large model size, which is about 1570 times larger than training solely on the algorithm task. This phenomenon reminds us of the \textit{emergent abilities} in Large Language Models~\citep{wei2022emergent}. The pretraining stage can also be seen as a multi-task learning process, where model has to remember numerous world knowledge while developing some general rules and abilities, such as reasoning. Our study suggests that this multi-task learning feature can be one important reason for the \textit{emergent abilities} in LLM.

Overall, we make three key contributions in this study, which are outlined as follows:
\begin{itemize}
    \item We introduce an innovative framework designed for analyzing and explaining the performance and training dynamics in consideration of both the size of the model and the quantity of training data.
    \item Utilizing this framework, we provide a nuanced illustration for the \textit{double descent} phenomenon and establish a predictive method for identifying instances of \textit{double descent} occurrence.
    \item By extending our framework to multi-task learning which consists of algorithm and pure memorization tasks, we convert algorithm tasks into an emergent ability. This offers a novel angle for understanding \textit{emergent abilities} in Large Language Models.
\end{itemize}

\textbf{Key Takeaways.} To better focus our readers' attention, we highlight the key takeaways from our analysis:
\begin{itemize}
    \item[\hyperlink{Takeaway-1}{1.}] Critical dataset size for generalization decreases with model size.
    \item[\hyperlink{Takeaway-2}{2.}] Memorization capacity increases with model size.
    \item[\hyperlink{Takeaway-3}{3.}] The training dynamics can be categorized into four types.
    \item[\hyperlink{Takeaway-4}{4.}] \textit{Progression} and \textit{grokking} differ in parameter norm variations despite both showing delayed generalization ability during training.
    \item[\hyperlink{Takeaway-5}{5.}] \textit{Double descent} is caused by training dynamics moving from \textit{progression} to \textit{ungrokking}, then to \textit{grokking}.
    \item[\hyperlink{Takeaway-6}{6.}] We can make \textit{double descent} more prominent by increasing the generalization difficulty.
    \item[\hyperlink{Takeaway-7}{7.}] \textit{Emergent ability} can be introduced by mixture of memorization tasks and generalization tasks.
    \item[\hyperlink{Takeaway-8}{8.}] Separating memorization and generalization in parameter space leads to faster emergence.
\end{itemize}

\section{Preliminary Study about Grokking}
\label{sec:study_about_grokking}


In this section, we first recall the preliminary setup of Grokking experiments in~\citet{DBLP:journals/corr/abs-2309-02390}, in which we can see the critical dataset size $D_{crit}^{M}$ as a function with model $M$ and extend the grokking experiments to different model sizes and training data size, resulting in an inverse relationship of critical dataset size $D_{crit}^{M}$ with model size, where larger models exhibit grokking with less training data.

\subsection{Experiments Setup}

\paragraph{Task} Following \citet{DBLP:journals/corr/abs-2201-02177} and \citet{DBLP:journals/corr/abs-2309-02390}, we conduct experiments on the modular addition task for generalization without specific illustration.
$$
(a + b)\ \mathrm{mod}\ P\ \mathrm{for}\ a,\ b\ \in (0, ... , P-1)\ \mathrm{and}\ P=113
$$
By using different tuples of $a$ and $b$, this task can be easily split into non-overlap train and validation set. Therefore, the memorization behaviour and generalization behaviour can also be distinguished by the validation performance. Although \citet{DBLP:conf/iclr/NandaCLSS23} propose a method to identify the generalization circuit of modular addition, we directly use validation performance to distinguish for simplicity. We use $D_{add}^{train}$ and $D_{add}^{val}$ to represent the data number of training set and validation set on this modular addition task.

\paragraph{Model} Following previous work~\citep{DBLP:conf/iclr/NandaCLSS23, DBLP:journals/corr/abs-2309-02390}, we train a 1-layer simplified decoder-only transformer~\citep{DBLP:conf/nips/VaswaniSPUJGKP17} model with 4 attention heads. We see the tasks above as classification tasks, where the label number is $P$. During training, a cross entropy loss $\mathcal{L}_{CE}$ with AdamW~\citep{DBLP:conf/iclr/LoshchilovH19} optimizer is utilized since the weight decay in AdamW has shown a significant effect for \textit{grokking}~\citep{DBLP:journals/corr/abs-2201-02177, DBLP:conf/iclr/NandaCLSS23, DBLP:journals/corr/abs-2309-02390}. We mainly vary model size by adjusting its hidden size dimension $d_h$.

\subsection{Grokking Experiments}

\begin{figure}
    \centering
    \includegraphics[width=0.95\linewidth]{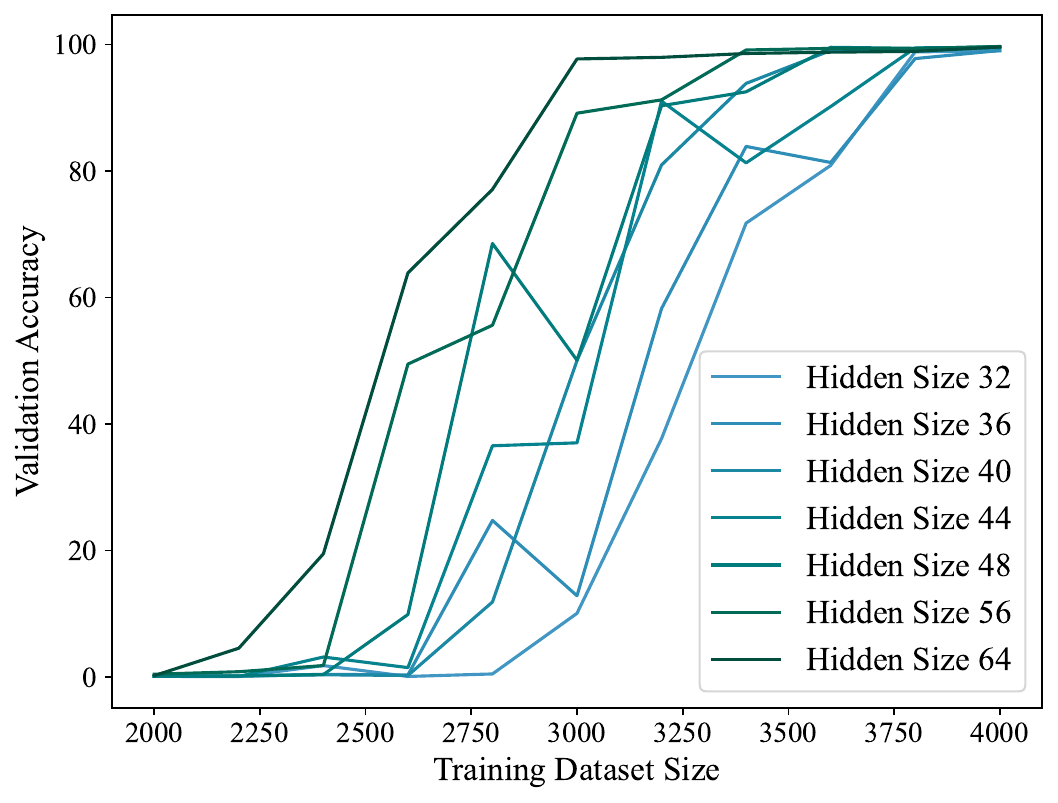}
    \caption{Final validation accuracy across various training dataset sizes and model hidden sizes. Larger models are represented in green, while smaller models are in blue. This figure demonstrates that models with larger hidden sizes attain near-perfect validation accuracy with comparatively less training data, indicating a reduced critical dataset size for these models.}
    \label{fig:critical_dataset_size}
\end{figure}

In this study, we explore the impact of varying training dataset sizes, denoted as $D_{add}^{train}$, which range from $2,000$ to $4,000$ in increments of $200$. Additionally, we investigate models with different hidden sizes, specifically $d_h \in \{32, 36, 40, 44, 48, 56, 64\}$. For each unique combination of dataset size and model hidden size, we perform experiments using 11 distinct random seeds, and their average performance is reported.

Our primary focus is to ascertain whether the modular addition task demonstrates the \textit{grokking} phenomenon in our experimental setup. Notably, all models quickly achieve perfect training accuracy within a few epochs, indicating their capability to efficiently memorize the training data. In Figure~\ref{fig:critical_dataset_size}, we present the final validation performance as a function of the training dataset size for different models. All of the combinations with near perfect validation accuracy show \textit{grokking} in our experiments, where validation accuracy grows up long after training accuracy being perfect.

\takeaway{Takeaway-1} From the analysis presented in Figure~\ref{fig:critical_dataset_size}, it is evident that models with larger hidden sizes tend to exhibit \textit{grokking} with smaller datasets. For instance, a model with a hidden size $d_h$ of $64$ achieves near-perfect validation accuracy with just $3,000$ training samples. In contrast, a model with a smaller hidden size of $32$ requires an increase in training data size to $4,000$ to achieve similar validation accuracy. This trend is consistent across models with intermediate hidden sizes ranging between $32$ and $64$.

In addition, our research also verifies the phenomenon termed \textit{semi-grokking}, as proposed by \citet{DBLP:journals/corr/abs-2309-02390}. This phenomenon occurs when a model exhibits partial generalization capabilities significantly after achieving perfect training accuracy. Specifically, \textit{semi-grokking} is observed when the number of training data points, $D_{add}^{train}$, closely matches the critical dataset size for the current model, denoted as $D_{crit}^{M}$. Under these conditions, the model's memorization and generalization circuits demonstrate comparable efficiencies. This balance prevents the model from fully transitioning from memorization to generalization, resulting in \textit{semi-grokking}.

For instance, as shown in Figure~\ref{fig:critical_dataset_size}, a model with a hidden size $d_h$ of $64$ exhibits \textit{semi-grokking} when the training dataset size, $D_{add}^{train}$, ranges between $2,000$ and $3,000$, where the critical dataset size, $D_{crit}^{M}$, for this model is approximately $3,000$. We observe that as the gap between $D_{add}^{train}$ and $D_{crit}^{M}$ narrows, there is a general improvement in the model's generalization ability, as reflected in its final validation accuracy. Furthermore, our experiments indicate that the \textit{semi-grokking} interval spans approximately $1,000$ training data points for all models under consideration.

\subsection{Memorization Experiments}
\label{subsec:memorization}
\begin{figure}[t]
    \centering
    \includegraphics[width=0.95\linewidth]{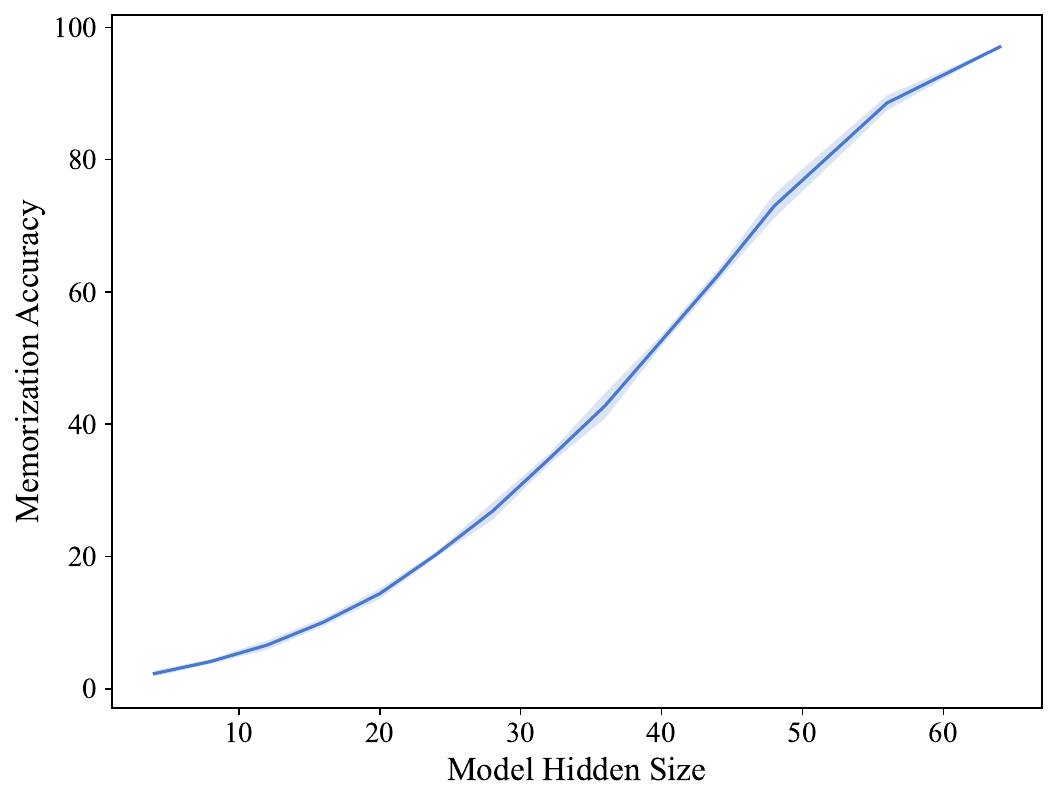}
    \caption{Graphical representation of the model's memorization capacity relative to its size. Each model is run using three distinct random seeds and the average performance is depicted. The light blue shaded region illustrates the 95\% confidence interval, which is notably narrow, highlighting the consistency of the memorization capacity across different model sizes.}
    \label{fig:memorization}
\end{figure}

\begin{figure}
    \centering
    \includegraphics[width=0.9\linewidth]{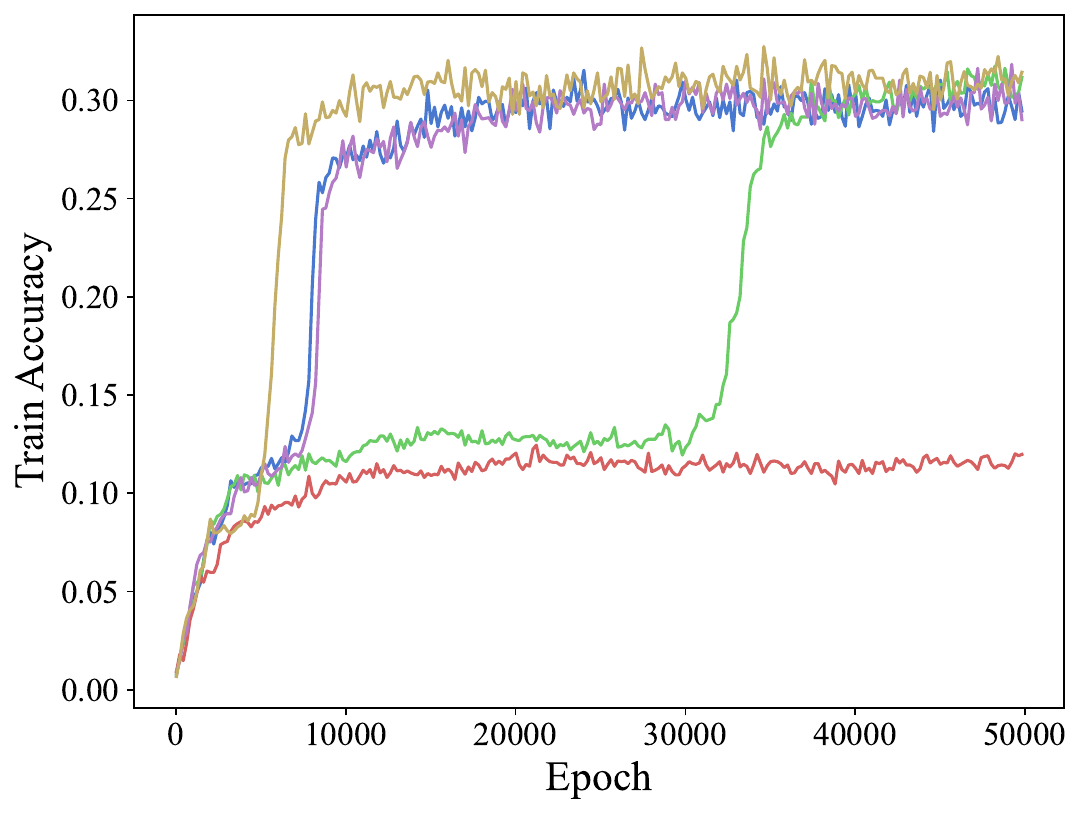}
    \caption{Progression experiments with $5$ different random seeds. All experiments are conducted with a training dataset size of $3000$ and a model hidden size of $8$. Curves with obvious bumps in training accuracy show \textit{progression} during training.}
    \label{fig:progression}
\end{figure}

Since \textit{grokking} requires the model to memorize all of the training data firstly, we designed experiments that specifically gauge the model's ability to memorize. For this purpose, we assign random labels to the training data~\citep{DBLP:journals/cacm/ZhangBHRV21}, compelling the model to focus solely on developing memorization circuits. Given the modular addition task's adherence to the commutative law, which could potentially affect the model's memorization capability, we assign identical random labels to $(a + b)\ \mathrm{mod}\ P$ and $(b + a)\ \mathrm{mod}\ P$. This approach ensures a more accurate assessment of the model's true memorization ability for this specific task. Our experiments encompass all $12,769$ pairs as training data, and we evaluate the model's accuracy on this entire set to represent its memorization capacity.

\takeaway{Takeaway-2} The findings, illustrated in Figure~\ref{fig:memorization}, reveal a distinctly positive correlation between the model size and its memorization capacity; larger models are capable of memorizing more training data. Additionally, it's noteworthy that the memorization capacity of the model remains relatively stable, as indicated by the small size of the 95\% confidence interval.

\section{Proposed Framework}
\label{sec:proposed_framework}

In this section, we introduce our framework designed to analyze both the training dynamics and final validation performance. This approach is grounded on two key assumptions, which are substantiated by the experiments detailed in Section~\ref{sec:study_about_grokking}.

\begin{assumption}
\label{ass:critical_dataset_size}
The critical dataset size for a model $M$ ($D_{crit}^{M}$) is negatively correlated to the model's size. This implies that larger models require less data to exhibit \textit{grokking}.
\end{assumption}

\begin{assumption}
\label{ass:memorization_capacity}
The memorization capacity of a model $M$ ($D_{mem}^{M}$) correlates positively with the model's size, indicating that larger models have a greater capacity to memorize training data.
\end{assumption}

\takeaway{Takeaway-3}\ Drawing from \cref{ass:critical_dataset_size} and \cref{ass:memorization_capacity}, we can construct a graphical representation for a specific task, as illustrated in Figure~\ref{fig:theory_graph}. This graph delineates the relationship between memorization capacity and critical dataset size, highlighting their intersection point (marked by a black star in Figure~\ref{fig:theory_graph}). Consequently, this demarcates four distinct zones, each representing a unique training dynamic. These dynamics will be discussed detailly in the subsequent discussion (Figure~\ref{fig:four_zones}).

\paragraph{Progression} In scenarios where the training dataset $D^{train}$ surpasses the memorization capacity $D_{mem}^M$ of a model $M$, the model is incapable of memorizing the entire dataset. This limitation leads to a two-stage learning dynamic. Initially, the model memorizes a portion of the training data, which does not translate into improved validation performance. Subsequently, the model begins to generalize, improving both its training accuracy and validation performance, as shown in Figure~\ref{fig:four_zones}~(a). However, this dynamic is not consistent across all runs and varies in the speed of generalization circuit development depending on different random seeds, as depicted in Figure~\ref{fig:progression}.

\paragraph{Ungrokking} When the training data $D^{train}$ falls below the memorization capacity $D_{mem}^M$ of the model, it can memorize the entire dataset. However, if $D^{train}$ is also less than the critical dataset size $D_{crit}^M$, memorization circuits outperform generalization circuits in efficiency. This leads the model to opt for pure memorization, resulting in negligible validation performance, as illustrated in Figure~\ref{fig:four_zones}~(b).

\paragraph{Grokking} In the situation where the training data $D^{train}$ is between the memorization capacity $D_{mem}^M$ and the critical dataset size $D_{crit}^M$, the model is able to memorize the entire training dataset. However, the circuits dedicated to memorization are less efficient compared to those for generalization. Initially, the model achieves perfect training accuracy through memorization. It then transitions to generalization circuits for higher efficiency, leading to a training dynamic demonstrated in Figure~\ref{fig:four_zones}~(c), where validation performance reaches near perfection well after the model has overfitted to the training set.

\paragraph{Semi-Grokking} In cases where the training data $D^{train}$ approximates the critical dataset size $D_{crit}^M$, the efficiency of memorization and generalization circuits becomes comparable. This parity causes the model to struggle with transitioning entirely to generalization circuits. Consequently, the final model is a combination of both memorization and generalization circuits, yielding moderate validation accuracy. Additionally, multiple plateaus in validation performance may be observed, indicating repeated shifts between memorization and generalization circuits, as shown in Figure~\ref{fig:four_zones}~(d).

\paragraph{Discussion} Despite \textit{progression} and \textit{grokking} show similar validation performance dynamic during training, it is crucial to note the distinct characteristics that differentiate \textit{progression} from \textit{grokking} in two fundamental aspects. Firstly, the generalization circuits in \textit{progression} emerge at a moderate level of training accuracy. In contrast, \textit{grokking} is characterized by the formation of generalization circuits only after the model has achieved perfect training accuracy. Secondly, the underlying mechanisms driving the formation of generalization circuits differ significantly between the two. In the case of \textit{progression}, generalization circuits are induced by the constraints of cross-entropy loss, as the model is unable to completely memorize all training data, preventing it from reaching near-zero training loss. Consequently, the generalization circuits in \textit{progression} are developed to minimize training loss while concurrently enhancing validation performance. On the other hand, the emergence of generalization circuits in \textit{grokking} is attributed to the model's preference towards more efficient circuits, particularly in terms of minimizing the model's parameter norm. (\takeaway{Takeaway-4}) This leads to a notable trend for generalization where, during \textit{grokking}, the model's parameter norm tends to decrease, whereas in \textit{progression}, an increase in the parameter norm is typically observed, which is shown in appendix~\ref{sec:further_progression}.

\section{Illustrate Double Descent}
\label{sec:doubel_descent}

\begin{figure*}
    \centering
    \includegraphics[width=1\linewidth]{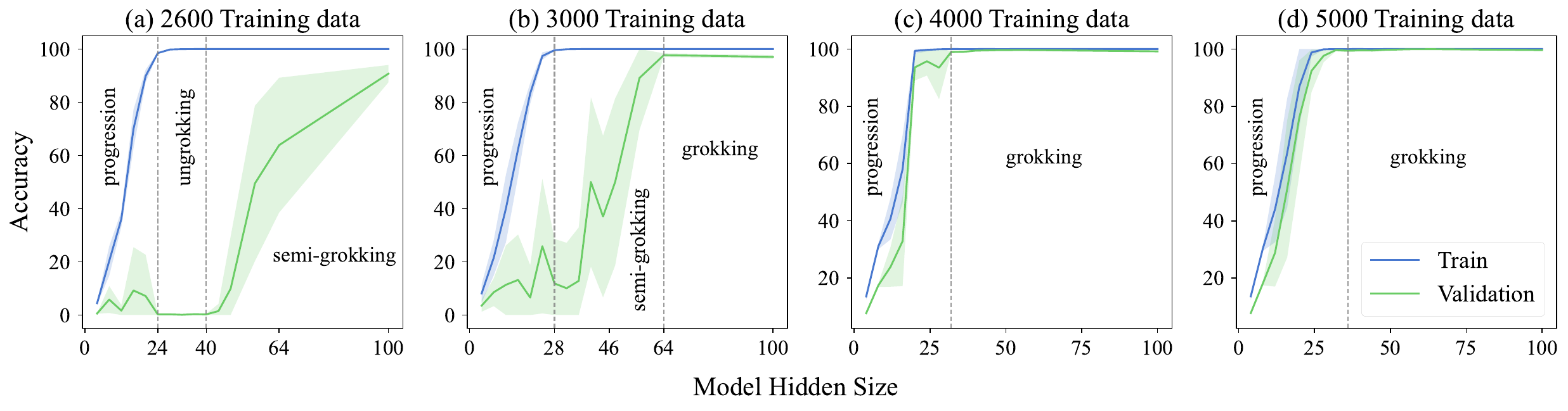}
    \caption{Experimental Results on the Modular Addition Task with Varying Training Data Sizes. Each experiment is conducted $11$ times using distinct random seeds, and the mean accuracy is reported. The light-colored regions denote $95\%$ confidence intervals. Consistent with our expectations, a smaller amount of training data tends to induce the \textit{double descent} phenomenon.}
    \label{fig:double_descent}
\end{figure*}

\begin{figure}
    \centering
    \includegraphics[width=0.9\linewidth]{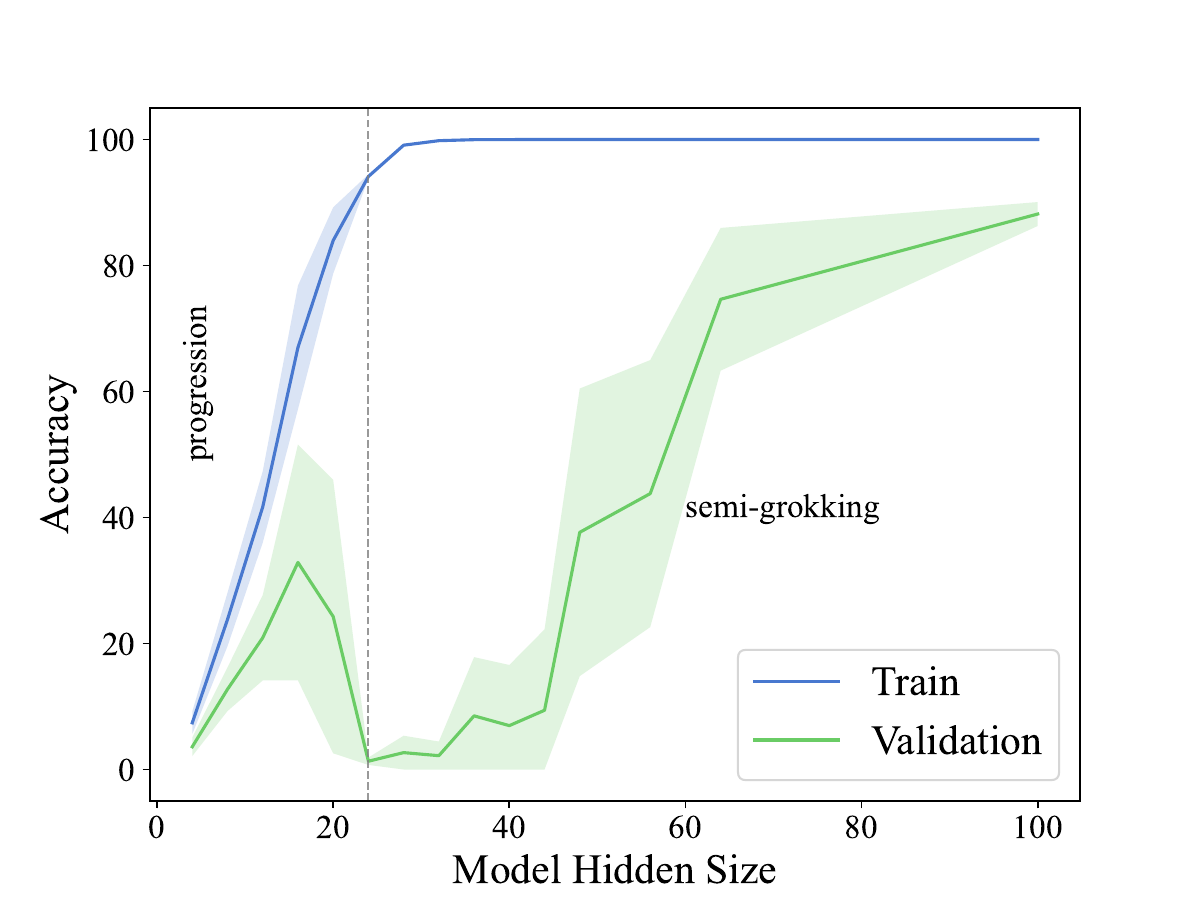}
    \caption{Experimental Analysis of $(a + b)^2\ \mathrm{mod}\ P$ with 3000 Training Data Points. When compared to Figure~\ref{fig:double_descent} (b), this graph illustrates that augmenting the generalization challenge leads to a more pronounced \textit{double descent} phenomenon with this quantity of training data.}
    \label{fig:polynomial3}
\end{figure}

The phenomenon of \textit{double descent}, as observed by \citet{DBLP:conf/iclr/NakkiranKBYBS20}, reveals an intriguing pattern wherein the increase in model size firstly detrimentally impacts validation performance before finally contribute positively to the validation performance. In this section, we provide a detailed exploration of the \textit{double descent} phenomenon, coupled with predictions regarding its occurrence given our framework in Section~\ref{sec:proposed_framework}. Subsequently, we undertake a series of experiments designed to validate our theoretical illustration.

\subsection{Illustration About Double Descent}
\takeaway{Takeaway-5}\ Our framework elucidates that the phenomenon of \textit{double descent} is likely to manifest when the number of training data, $D^{train}$, is substantially lower than the intersection point of two curves in Figure~\ref{fig:theory_graph}. In such scenarios, the model undergoes a series of stages: \textit{progression}, \textit{ungrokking}, \textit{semi-grokking}, and finally \textit{grokking}. Initially, as the model size increases during the \textit{progression} phase, there is an enhancement in its generalization ability. However, this ability declines, potentially to zero, upon transitioning into the \textit{ungrokking} stage. This decline is in alignment with the critical interval of model size postulated by \citet{DBLP:conf/iclr/NakkiranKBYBS20}. As the model continues to grow, it enters the \textit{semi-grokking} and \textit{grokking} phases, where its generalization ability is revived, leading to a secondary increase. This relationship results in the \textit{double descent} curve observed in validation performance\footnote{The ``descent'' in double descent refers to validation error/loss, therefore in terms of validation performance, it's validated by the two ``increase'' stage.}. Conversely, in instances where the training data exceeds the intersection point, the model predominantly experiences \textit{progression} and \textit{grokking}, resulting in a consistent upsurge in generalization ability as the model size increases. Under these conditions, the \textit{double descent} phenomenon does not occur.

\subsection{Experiments}

To validate our illustration, we carried out a series of experiments centered on the modular addition task, as detailed in Section~\ref{sec:study_about_grokking}. We set the training data sizes to $\{2600, 3000, 4000, 5000\}$ and varied the model's hidden size from $4$ to $100$, following the sequence $\{4, 8, 12, 16, 20, 24, 28, 32, 36, 40, 44, 48, 56, 64, 100\}$. Each configuration was tested across $11$ experiments with distinct random seeds, and the average performance was reported, accompanied by a $95\%$ confidence interval. These results are depicted in Figure~\ref{fig:double_descent}.

Observations from Figure~\ref{fig:double_descent} (a) reveal that with a training dataset of $2600$ samples, the models transition through the stages of \textit{progression}, \textit{ungrokking}, \textit{semi-grokking}\footnote{Due to the small number of training data, \textit{grokking} doesn't happen in this model size range.}. In the \textit{progression} phases, larger models generally demonstrate improved validation performance. However, the \textit{ungrokking} phase, evident when the model's hidden size ranges between $24$ and $40$, leads to zero validation performance, resulting in the \textit{double descent} pattern for this dataset size.

Increasing the training data slightly to $3000$ samples eliminates the \textit{ungrokking} stage. The models then progress through \textit{progression}, \textit{semi-grokking}, and \textit{grokking}, resulting in a less pronounced \textit{double descent} curve. Still, a dip in validation performance is noticeable for model sizes between $28$ and $36$, as shown in Figure~\ref{fig:double_descent} (b). Outside this range, there is a general trend of increasing performance with larger model sizes.

With the training data further increased to $4000$ and $5000$, models bypass the \textit{ungrokking} and \textit{semi-grokking} stages, transitioning directly from \textit{progression} to \textit{grokking}. This leads to a consistent improvement in validation performance as model size increases, without any occurrence of \textit{double descent}, as illustrated in Figure~\ref{fig:double_descent} (c) and (d).

Beyond \textit{double descent}, a phenomenon termed \textit{model-wise grokking} by \citet{davies2022unifying} is also observed: smaller training datasets result in validation accuracy reaching near perfect long after training accuracy in terms of model size.  However, this phenomenon also disappears with increased training data, as models are able to achieve perfect training and validation performance during the \textit{progression} stage with these larger datasets.

\subsection{Make Double Descent More Prominent}
\takeaway{Takeaway-6}\ Under our proposed framework, elevating the curve of critical dataset size causes a shift in the intersection point towards the upper right. This results in an expanded range of training data sizes that exhibit the \textit{double descent} phenomenon. Consequently, a quantity of training data that previously did not demonstrate \textit{double descent} can be transformed to one that does. To move the critical dataset size curve to the upper right, what we need is to increase the complexity involved in formulating generalization circuits for the task.

To further validate our hypothesis, we designed experiments centered around a task more complex than the modular addition task. This task is defined as:
$$
(a + b)^2\ \mathrm{mod}\ P\ \mathrm{for}\ a,\ b\ \in (0, ... , P-1)\ \mathrm{and}\ P=113
$$
Given that this task also adheres to the commutative law, it does not affect the memorization capacity curve\footnote{It is easier for model to memorize training data which obeys commutative law, since it only needs to memorize half of the training data.}. For a direct comparison, we used a training dataset of 3000 instances, the same size that did not exhibit a clear \textit{double descent} in the modular addition task. Each experimental setting was replicated 11 times using distinct random seeds. The outcomes of these experiments are depicted in Figure~\ref{fig:polynomial3}.

The results from this more complex task reveal a more pronounced \textit{double descent} phenomenon in the validation performance compared to the one observed in Figure~\ref{fig:double_descent} (b), which utilized an identical quantity of training data. Notably, in this task, the model exhibited a zero validation performance when the hidden size of the model was set to 24. This outcome was not observed in the modular addition task with the same training data size of $3000$. These behaviours corroborate our previously stated hypothesis.

\subsection{Discussion}
The analysis of the \textit{progression} stage reveals varying relationships between the model's validation performance and its size. For smaller training datasets that exhibit the \textit{double descent} phenomenon, the validation performance typically follows an increasing-then-decreasing trend. This trend smoothly transitions into the zero validation performance characteristic of the \textit{ungrokking} stage. Conversely, with larger training datasets, the validation performance consistently improves alongside increases in model size, smoothly transitioning into the near-perfect validation performance observed during the \textit{grokking} stage. This variation can be attributed to the interplay between cross-entropy loss and the model's tendency to favour more efficient methods. As models near the \textit{ungrokking} stage from \textit{progression}, cross-entropy loss drives the model towards generalization for lower training loss, but the efficiency of memorization circuits pulls the model towards pure memorization. This conflict results in a decreasing trend in validation performance in the latter stage of \textit{progression}. However, as models transition from \textit{progression} to \textit{grokking}, both cross-entropy loss and efficiency considerations align, leading to an uninterrupted improvement in validation performance.

\section{Multi-Task Learning Leads to Emergent Ability}
\label{sec:emergent_ability}
\begin{figure}
    \centering
    \includegraphics[width=0.95\linewidth]{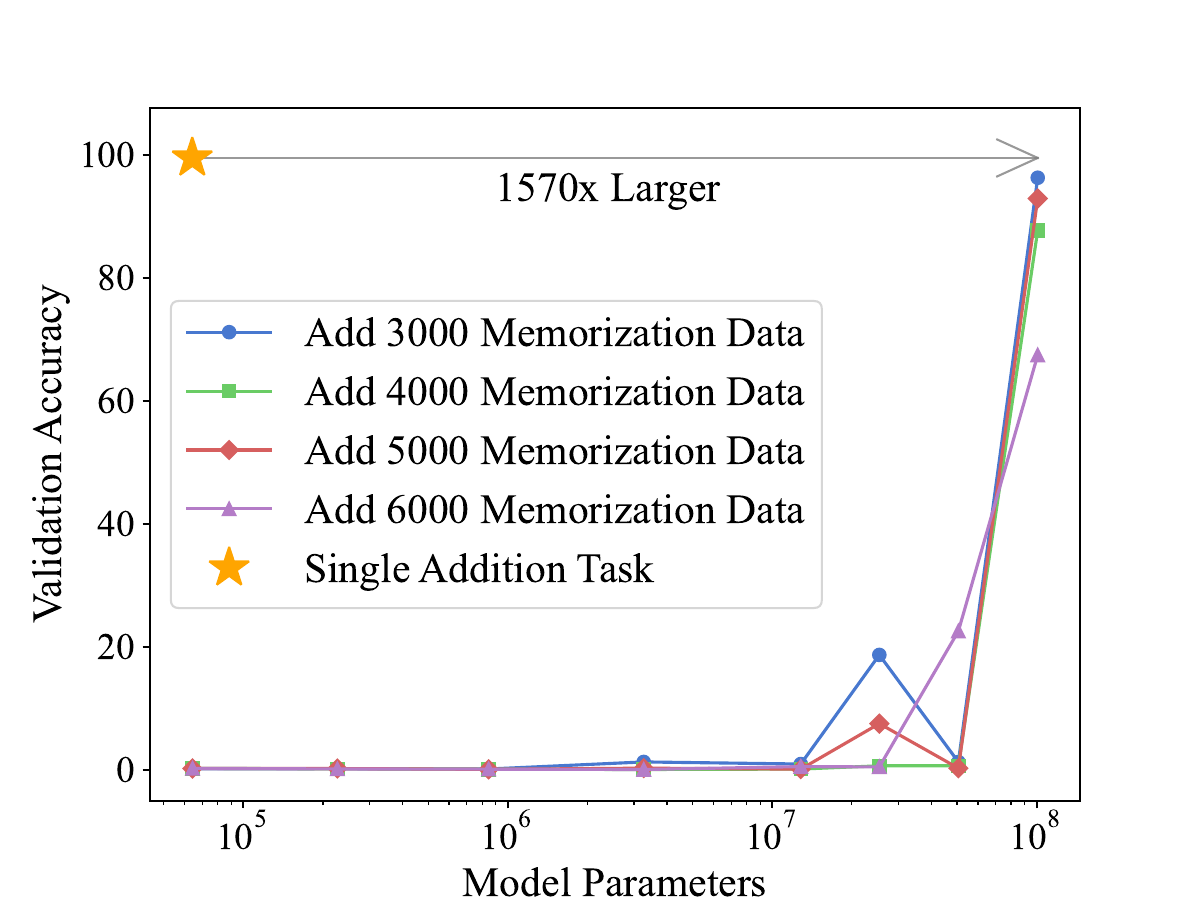}
    \caption{Adding a pure memorization task into the modular addition task makes it become an emergent ability.}
    \label{fig:emergent_ability}
\end{figure}

In this section, we expand our research to the multi-task learning paradigm, which combines an algorithm task, such as modular addition, with a task focused solely on memorization. This approach reveals that the model's generalization ability on the algorithm task remains negligible until the model reaches a substantially larger size—specifically, 1570 times larger than training on a single task. As a result, the validation performance on algorithm task show an emergent phenomenon relative to model size.

\paragraph{Experiment Setup} Our experiments utilize the modular addition task as the algorithm component. For the memorization task, we assign random labels to a subtraction task, compelling the model to memorize these associations. By incorporating different calculation symbols ($+, -$) into the input tokens, we ensure that the memorization and algorithm task have no overlapping inputs. Our experiments involve $3000$ data points from the modular addition task and a varying number of memorization data points, ranging from $3000$ to $6000$. We adjust the model size from a 1-layer transformer with a hidden size of 64 to an 8-layer transformer with a hidden size of 1024. Each experiment is conducted three times, and we report the highest validation accuracy for each configuration to showcase each model's optimal potential. The results are presented in Figure~\ref{fig:emergent_ability}.

\takeaway{Takeaway-7}\ We observe that incorporating pure memorization data significantly impedes smaller models in developing generalization circuits. The emergence of generalization ability in the modular addition task is notable, with models typically displaying substantial validation performance at relatively larger sizes, approximately 1570 times larger than those trained solely on the modular addition task. Additionally, the volume of memorization data appears to have minimal impact on the emergent model size.

\paragraph{Discussion} From the perspective of the competition between memorization and generalization circuits, the presence of a pure memorization task prevents the model from transitioning from memorization to generalization after mastering all training data, as there are no generalization circuits for the pure memorization task. However, once a sufficiently large model size is attained, the model's memorization capacity significantly exceeds the training data volume, allowing it to allocate extra parameters for generalization in the algorithm task. The experiment in appendix~\ref{sec:further_emergent}, which reduces the emergent model size by allocating separate parameters for algorithm and memorization data, further supports this hypothesis. This phenomenon echoes the \textit{emergent abilities} observed in current Large Language Models. Considering that the pretraining stage resembles a multi-task learning scenario—where the model must retain a vast array of world knowledge while acquiring general rules and capabilities, such as in context learning~\citep{NEURIPS2020_1457c0d6} and multi-step reasoning~\citep{NEURIPS2022_9d560961}—our experiments may provide fresh insights into the \textit{emergent abilities} in Large Language Models. This observation further elucidates the hypothesis proposed by \citet{hu2023predicting}, where it is hypothesized that emergent abilities are formed through the competition of different neural circuits.

\section{Related Work}
\label{sec:related_work}

\paragraph{Grokking} The phenomenon of \textit{grokking}, where models demonstrate exceptional generalization capabilities well beyond the point of overfitting to training data, was first identified by \citet{DBLP:journals/corr/abs-2201-02177} in various algorithm tasks. \citet{DBLP:journals/corr/abs-2206-04817} demonstrate that \textit{grokking} often comes with "Slingshot Effects," which may play a pivotal role in its emergence. Delving into the underlying mechanisms, \citet{DBLP:conf/nips/LiuKNMTW22} approached \textit{grokking} from a representation learning standpoint, uncovering its association with structured representation development. Further, \citet{DBLP:conf/iclr/LiuMT23} discovered that grokking is not confined to algorithm tasks but also manifests in a broader spectrum of realistic tasks, given specific initial conditions. Additionally, \citet{DBLP:journals/corr/abs-2306-13253} explored predictive markers of grokking, highlighting "oscillations" within the loss landscape as potential indicators. A novel perspective by \citet{DBLP:journals/corr/abs-2309-02390} suggests that grokking can be interpreted through the competition between memorization and generalization circuits, influenced by the efficiency of these circuits. This demonstration can be included in our framework by a vertical line on the right of the intersection point in Figure~\ref{fig:theory_graph}.

\paragraph{Double Descent} The concept of \textit{double descent}, as introduced by \citet{Belkin_2019}, illustrates a unique pattern in model validation performance: an initial increase, followed by a decrease, and then a subsequent increase, in correlation with the growing size of the model. This dip in performance typically coincides with the model’s training error nearing zero. Expanding upon this concept, \citet{DBLP:conf/iclr/NakkiranKBYBS20} conducted a comprehensive examination of the double descent phenomenon across varying model architectures, datasets, and optimization techniques. In a parallel effort, \citet{davies2022unifying} attempted to bridge the concepts of \textit{double descent} and \textit{grokking}. This was approached through a proposed duality between model size and scaling time, though this hypothesis remains to be empirically verified.

\paragraph{Emergent Abilities} The concept of \textit{emergent abilities} has garnered significant interest in the era of the development of Large Language Models (LLMs). \citet{wei2022emergent} provide a thorough investigation into different abilities across various models, characterizing them as capabilities that are absent in smaller models but present in larger ones. \citet{DBLP:conf/iclr/CaballeroGRK23} introduced a complex function, modeled on a piece-wise power law, to encapsulate various phenomena, including \textit{emergent abilities}. \citet{schaeffer2023are} attributed the emergence of these abilities to the non-smooth metrics employed in task evaluation. Taking a predictive angle, \citet{hu2023predicting} succeeded in forecasting certain \textit{emergent abilities} by employing a metric with infinite resolution. Different from these works, our research delves into this phenomenon from a unique angle, examining the competition between memorization and generalization circuits within neural models.

\section{Conclusion}
\label{sec:conclusion}

In this paper, we introduce an innovative framework designed to analyze varying training dynamics across different model sizes and training dataset quantities. This framework is grounded in the competition between memorization and generalization circuits. Leveraging this framework, we offer a comprehensive illustration of \textit{double descent} and validate our predictions regarding its occurrence under two distinct scenarios. By integrating memorization and generalization tasks, we successfully induce emergent behaviour in generalization tasks, shedding new light on the understanding of \textit{emergent abilities}. A notable limitation of our study, however, is its exclusive focus on algorithm tasks. Expanding this research to encompass more realistic tasks and models in future work will be crucial for a deeper and more comprehensive understanding of deep learning mechanisms.

\section{Impact Statement}
This paper presents work aimed at advancing the field of mechanistic understanding and training dynamics of large language models. As our work is purely theoretical, we acknowledge the potential societal consequences but do not think of any specific issues need to be highlighted in this context.



\bibliography{example_paper}
\bibliographystyle{icml2024}

\newpage
\appendix
\onecolumn
\section{Parameter Norm Variations on Progression and Grokking}
\label{sec:further_progression}

\begin{figure}[h]
    \centering
    \includegraphics[width=0.95\linewidth]{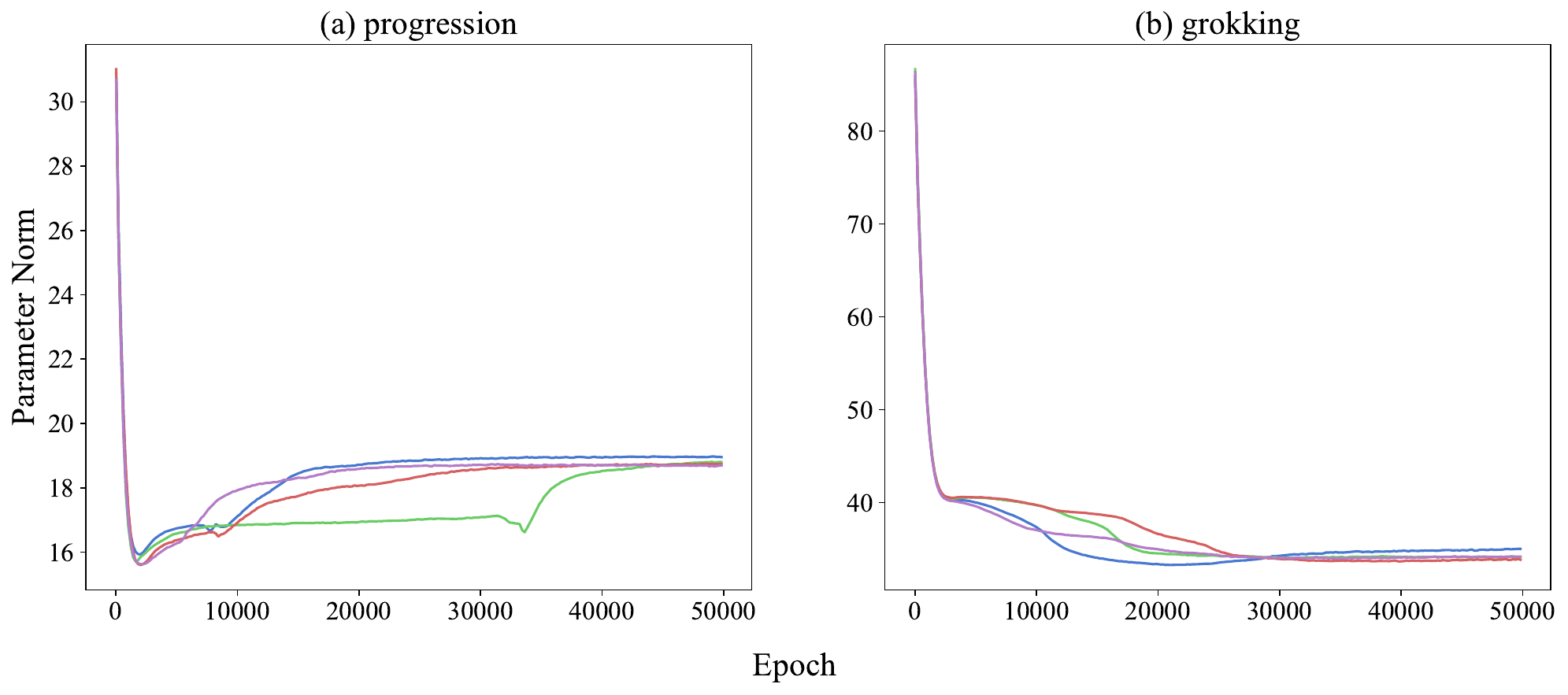}
    \caption{Variation in parameter norm during training for \textit{progression} and \textit{grokking} with varied seeds. For \textit{progression}, experiments are performed using $3000$ training data points and a model hidden size of $8$. In contrast, \textit{grokking} experiments utilized the same number of training data but with a model hidden size of $64$.   Notably, the parameter norm in \textit{progression} consistently exhibits a marked increase, as exemplified by the green line between 30,000 and 40,000 epochs,  with the transition to generalization. Conversely, in \textit{grokking}, the parameter norm demonstrates a decreasing trend as generalization circuits become established, highlighting the distinct mechanisms driving generalization in each case.}
    \label{fig:norm}
\end{figure}

As we have discussed in Section~\ref{sec:proposed_framework}, the concepts of \textit{progression} and \textit{grokking} differ in two fundamental ways. The first difference lies in the training accuracy at the point of generalization, a distinction that is clearly observable in Figure~\ref{fig:four_zones}. The second difference pertains to the underlying reasons for the emergence of generalization circuits. In the case of \textit{progression}, generalization is triggered by non-zero training loss in memorization circuits, leading to a reduction in both training and validation losses by generalization. Conversely, \textit{grokking} occurs when generalization circuits prove to be more efficient than those used for memorization, a phenomenon that is evident from the changing parameter norm of the model. This distinction is highlighted by the observed trends in parameter norms: during the shift from memorization to generalization, \textit{grokking} exhibits a decreasing parameter norm, whereas \textit{progression} demonstrates an increasing parameter norm, as depicted in Figure~\ref{fig:norm}.

\section{Further Experiments on Emergent Ability}
\label{sec:further_emergent}

\begin{figure}[ht]
    \centering
    \includegraphics[width=0.6\linewidth]{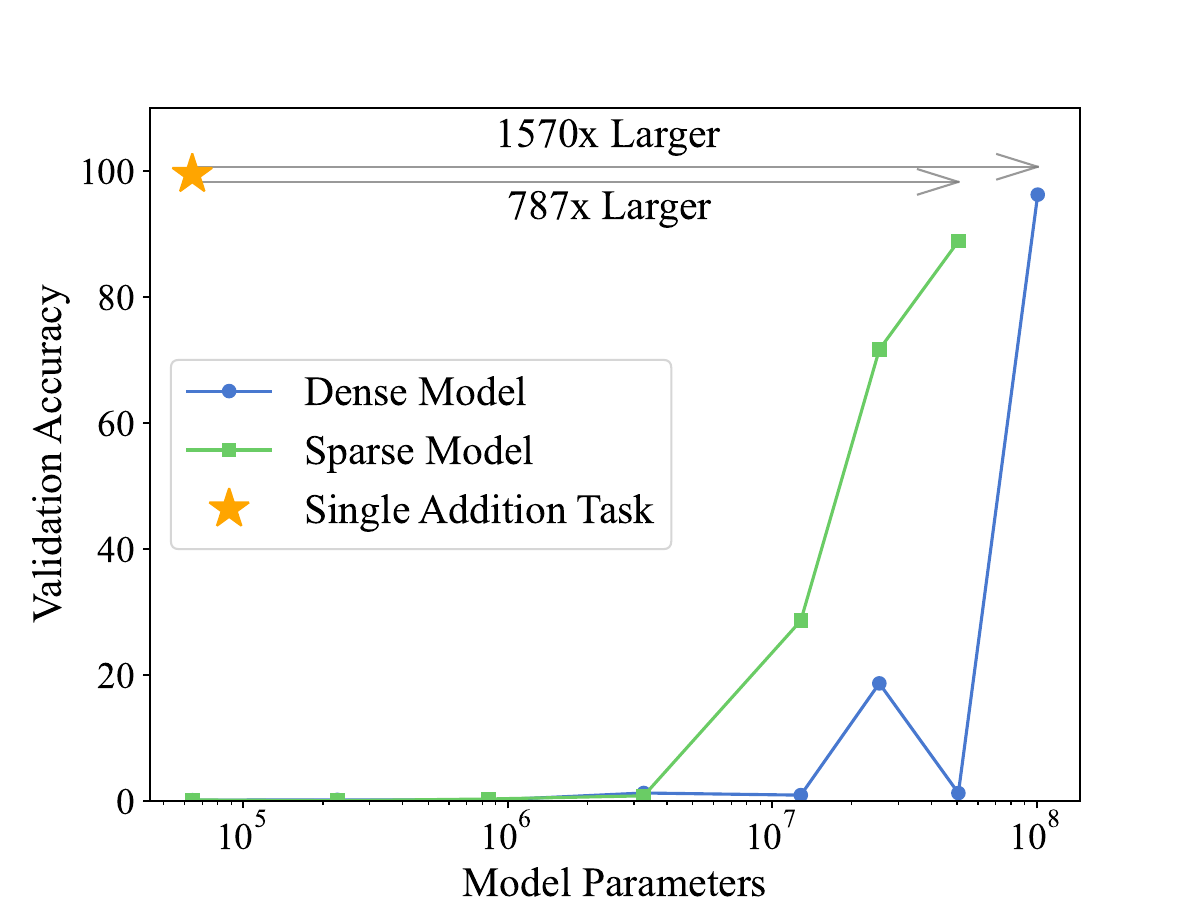}
    \caption{Experiments on multi-task learning with a pure memorization task and modular addition task. This study delves into the effects of manually constructing a sparse model to separately address pure memorization and modular addition data. Through this approach, we significantly accelerate the emergence of the modular addition capability in terms of model size. This finding highlights the crucial role of functional differentiation within neural models in fostering the emergence of new abilities.}
    \label{fig:emergent_ability_moe}
\end{figure}

\takeaway{Takeaway-8}\ To enhance our understanding of how combining a pure memorization task with a modular addition task results in emergent ability, we hypothesize that the inclusion of a pure memorization task inhibits the model's shift from memorization to generalization. This occurs because there are no generalization circuits specifically for the pure memorization task. To investigate this further, we designed experiments to segregate parameters dedicated to memorization and generalization. Prior studies suggest that the feed-forward layer in the Transformer architecture predominantly facilitates memorization~\citep{DBLP:conf/emnlp/GevaSBL21}. Additionally, this ffn layer in transformer is also crucial for the generalization process in the modular addition task~\citep{DBLP:conf/iclr/NandaCLSS23, DBLP:conf/icml/ChughtaiCN23}. Consequently, we partitioned the feed-forward layer into two specialized sections by dividing the intermediate dimension, akin to the current MoE architecture~\citep{DBLP:journals/jmlr/FedusZS22}. In this setup, one section exclusively processes the modular addition task data, while the other focuses solely on the memorization task data. Our experiments, conducted on $3000$ instances each of modular addition and memorization data, are depicted in Figure~\ref{fig:emergent_ability_moe}. The results demonstrate that manually creating a sparse network significantly accelerates emergence of the modular addition task capabilities with the same training dataset. This finding underscores the critical role of functional differentiation in neural models for the development of diverse abilities.

\end{document}